\documentclass[review]{elsarticle}

\usepackage{enumerate}
\usepackage{times}
\usepackage{algorithm}
\usepackage{algorithmic}
\usepackage{epstopdf}
\usepackage{marvosym}
\usepackage{url}
\usepackage{amsfonts}
\usepackage{nicefrac}
\usepackage{microtype}
\usepackage{bm}
\usepackage{makecell}
\usepackage{color}
\usepackage{caption}
\usepackage{graphicx, subfigure, booktabs, amsmath, amssymb, epsfig, hyperref}
\usepackage{amsthm}

\usepackage{diagbox}
\usepackage{multirow}
\usepackage{array}
\usepackage{amsmath}
\usepackage{amssymb}
\usepackage{dsfont}
\usepackage[normalem]{ulem}
\useunder{\uline}{\ul}{}

\journal{Information Fusion}

\begin{document}

\begin{frontmatter}

\title{MST-GAT: A Multimodal Spatial-Temporal Graph Attention Network for Time Series Anomaly Detection}

\hypersetup{pdfauthor={Name}}

\author{Chaoyue Ding}
\ead{cydingcs@gmail.com}
\author{Shiliang Sun}
\author{Jing Zhao}

\address{School of Computer Science\label{ecnu} and Technology, East China Normal University, \\3663 North Zhongshan Road, Shanghai 200062, China}

\begin{abstract}
Multimodal time series (MTS) anomaly detection is crucial for maintaining the safety and stability of working devices (e.g., water treatment system and spacecraft), whose data are characterized by multivariate time series with diverse modalities. Although recent deep learning methods show great potential in anomaly detection, they do not explicitly capture spatial-temporal relationships between univariate time series of different modalities, resulting in more false negatives and false positives. In this paper, we propose a multimodal spatial-temporal graph attention network (MST-GAT) to tackle this problem. MST-GAT first employs a multimodal graph attention network (M-GAT) and a temporal convolution network to capture the spatial-temporal correlation in multimodal time series. Specifically, M-GAT uses a multi-head attention module and two relational attention modules (i.e., intra- and inter-modal attention) to model modal correlations explicitly. Furthermore, MST-GAT optimizes the reconstruction and prediction modules simultaneously. Experimental results on four multimodal benchmarks demonstrate that MST-GAT outperforms the state-of-the-art baselines. Further analysis indicates that MST-GAT strengthens the interpretability of detected anomalies by locating the most anomalous univariate time series.
\end{abstract}

\begin{keyword}
Multimodal time series \sep anomaly detection \sep graph attention networks \sep unsupervised learning
\end{keyword}

\end{frontmatter}

\section{Introduction}
\label{sec:intro}

Anomaly detection has gained much attention in different fields (e.g., images~\cite{vojir2021road}, text~\cite{ruff2019self}, time series~\cite{ren2019time}, etc.), aiming to find instances that deviate significantly from all other observations~\cite{chandola2009anomaly,chalapathy2019deep}.
In this paper, we focus on multimodal time series (MTS) anomaly detection, which is the subtask of anomaly detection.
MTS anomaly detection is commonly used to monitor diverse modalities (e.g., temperature, speed, and power) of sensors in industrial devices and information technology systems (i.e., entities), and the data stream from each sensor is seen as a univariate time series.
Multimodal time series data can facilitate the detection of complex anomalies, which is not obvious when monitoring each modality independently ~\cite{park2016multimodal}.
Furthermore, before the entity is partially or fully disrupted, the timely detection of anomalies helps the user to troubleshoot.

Conventionally, experienced engineers manually create static thresholds for each monitored time series to perform anomaly detection.
However, with the exponential growth of data size in recent years, this manual approach will be labor-intensive~\cite{audibert2020usad}.
Moreover, determining an optimal threshold for each sensor is challenging, especially when multimodal sensing is employed in the entity.
Many anomaly detection methods have been proposed to solve the above problem, and they combine the anomaly detection results of all univariate time series in an entity to detect anomalies~\cite{ren2019time,li2021multivariate}.
However, an entity of multimodal time series often involves massively interconnected univariate time series, which continuously generate multimodal time series data, and these sensor data are typically correlated in complex non-linear ways.
Thus, a single univariate time series fails to respond to the overall state of the entity, and methods that simply combine the detection results of multiple univariate time series tend to perform poorly.
MTS anomaly detection has been challenging due to the complex spatial dependence (e.g., topological structure and modal correlation) and temporal dependence (e.g., period and trend) of multimodal time series. Besides, multimodal time series contains not only correlations between time series from the same modality (referred to as intra-modal correlations) but also correlations between time series from different modalities (referred to as inter-modal correlations).

Previous methods for MTS anomaly detection take temporal dependence into account, including support vector regression~\cite{kromanis2013support}, Bayesian models~\cite{hill2007real}, autoregressive integrated moving average (ARIMA)~\cite{zhang2003time} and recurrent neural network (RNN)-based models~\cite{shipmon2017time}. These methods can capture the dynamic changes in temporal dimension but ignore the spatial dependence between different time series. To remedy this, some researchers introduced convolutional neural networks (CNNs) to better model spatial relationships~\cite{ren2019time}. However, CNNs are usually applied to regular data such as image, video and speech data, which would lead to inferior performance in graph data due to the complex topology among multimodal time series. Graph neural networks (GNNs)~\cite{wu2020comprehensive} are more effective paradigms to build complex topological relationships in graph data, which have also been developed for anomaly detection and achieved promising results. Specifically, Hang et al.~\cite{zhao2020multivariate} adopted graph neural networks and gated recurrent unit (GRU)~\cite{chung2014empirical} to study the spatial-temporal relationships of time series. Although the previous methods have made fruitful progress, they fail to explicitly capture the multimodal correlation among multimodal time series.

Another challenge is to provide interpretation for anomaly detection results.
To provide users with more valuable information, MST-GAT interprets each anomaly by locating a univariate time series that is most likely to cause this anomaly.
The reconstruction probability of the reconstruction-based method is usually used to interpret detected anomalies.
For instance, OmniAnomaly~\cite{su2019robust} leveraged variational autoencoder (VAE)~\cite{kingma2013auto} to produce reconstruction probabilities for each time series, which is used to explain the detected anomalies.
Although these methods can capture the stochasticity of entire time series, researches show that they fail to perform well on periodic scenarios, while prediction-based methods can overcome this deficiency~\cite{zhao2020multivariate}.

In this paper, we propose the multimodal spatial-temporal graph attention network, termed MST-GAT, which adopts the prevalent graph attention networks (GATs)~\cite{velivckovic2018graph} to explicitly capture modal dependencies between multimodal time series.
More specifically, we design the multimodal graph attention network (M-GAT), which includes a multi-head attention module and two relational attention modules, i.e., intra- and inter-modal attention, to capture the spatial dependencies between multimodal time series. Explicitly modeling different relationships in multimodal time series is conducive to obtaining a better feature representation of input data.
We then introduce a temporal convolution network to capture temporal dependencies in each time series with a standard convolution operation on time slices.
Additionally, we jointly optimize a reconstruction module and a prediction module to integrate their advantages.
The reconstruction module accounts for reconstructing the input data, while the prediction module aims at predicting the feature of the next timestamp.
The reconstruction probability and the prediction error are further used to explain the detected anomalies.

The main contributions of this paper are summarized as follows:
\begin{itemize}
\item We propose MST-GAT, a novel MTS anomaly detection method based on graph attention networks. To the best of our knowledge, MST-GAT pioneers the exploration of explicitly modeling spatial-temporal dependencies in multimodal time series data for anomaly detection.
\item We jointly optimize a variational autoencoder-based reconstruction module and a multilayer perceptron (MLP)-based prediction module to integrate their advantages. MST-GAT achieves the highest F1-score, all above $0.60$, the best AUC, all above $0.92$, outperforming the strong baselines on benchmark datasets. Ablation studies further prove the effectiveness of the different modules in the MST-GAT.
\item We devise an efficient anomaly interpretation method for MTS anomaly detection based on the reconstruction and prediction results. MST-GAT is well interpretable and is capable of obtaining results that are consistent with human intuition.
\end{itemize}

{ The rest of the paper is structured as follows. Section 2 presents the related works. Section 3 introduces the proposed MST-GAT. In Section 4, experimental results demonstrate the effectiveness of the proposed method. Finally, conclusions and future research directions are given in the last section.}

\section{Related Works}
{
This section briefly introduces related works, including time series anomaly detection, graph neural networks, and multimodal machine learning. 
}
\subsection{Time Series Anomaly Detection}
Time series anomaly detection has been investigated for decades, and various types of approaches have been proposed~\cite{braei2020anomaly,erhan2021smart}.
Traditional anomaly detection approaches can be classified into clustering~\cite{kiss2014data}, distance-based~\cite{chaovalitwongse2007time}, density-based~\cite{ma2003time}, and isolation-based~\cite{puggini2018enhanced} methods.
Recently, deep learning approaches have attracted much attention due to the powerful representational capabilities of deep neural networks~\cite{zhang2019deep}.
Existing deep learning methods can be categorized into two paradigms, namely reconstruction-based and prediction-based methods.

A reconstruction-based method learns the potential distribution of the entire time series.
Deep autoencoding Gaussian model (DAGMM)~\cite{dagmm} obtained low-dimensional features and reconstruction-based anomaly scores by combining a deep autoencoder network and a Gaussian mixture model (GMM).
OmniAnomaly~\cite{su2019robust} employed VAE into an end-to-end structure to reconstruct the input data, and it detected anomalies according to the reconstruction probability.
RAMED~\cite{shen2021time} utilized a multi-resolution network to encourage reconstructed outputs to match the global temporal shape of input.
A prediction-based method predicts the value of the following timestamp and anomalies according to the prediction residual.
Hundman et al.~\cite{hundman2018detecting} demonstrated the feasibility of long short-term memory (LSTM) in detecting spacecraft anomalies and introduced an approach for setting thresholds dynamically without relying on annotations.
Graph deviation network (GDN)~\cite{deng2021graph} utilized graph attention networks (GATs) to perform structure learning in multivariate time series and interpreted a detected anomaly by attention weights.

Previous works have proved that reconstruction-based and prediction-based methods are complementary in different scenarios~\cite{zhao2020multivariate}. Therefore, we propose a joint network to integrate the advantages of these two paradigms. Nevertheless, none of the existing methods consider explicitly capturing the relationship between multimodal data.
Our work aims to address this problem by using multimodal graph attention to explicitly construct spatial-temporal dependencies within multimodal time series.

\subsection{Graph Neural Networks}
Graph neural networks (GNNs) have gained remarkable success in graph structure data such as social networks~\cite{yuan2019graph} and medical science~\cite{wang2021covid}.
Typical GNNs supposes that the representation of a node is affected by its neighboring nodes in a graph structure.
Graph convolutional networks (GCNs) include spectral methods and spatial methods~\cite{kipf2016semi}.
Spectral methods suffer from the drawback of basic dependence, and spatial methods are limited by the lack of shift-invariance~\cite{duvenaud2015convolutional,monti2017geometric}.

Attention mechanisms have become an effective and widely used component of the sequence-to-sequence models in many deep learning applications~\cite{bahdanau2014neural,vaswani2017attention}.
Recently, attention mechanisms have been introduced to graph neural networks.
Graph attention networks (GATs)~\cite{velivckovic2018graph} utilized the attention mechanisms to assign aggregation weights to neighboring nodes.
Relevant variants of graph attention networks have made progress in tasks related to time series modeling, e.g., traffic flow forecasting~\cite{zhang2020spatial} and time series forecasting~\cite{cirstea2021graph}.
Graph attention networks can better extract the spatial feature and shows superior performances than graph convolutional neural network in the directed graph~\cite{zhu2021dyadic}.
GATs map the input feature vectors ${h}$ into the aggregated representation ${h^{\prime}}$ using the attention mechanisms.
The attention score $a_{ij}$ is formulated as:
\begin{gather}
    a_{i j}=\frac{\exp \left(\hat{a}_{i j}\right)}{\sum_{k \in \mathcal{N}_{i}} \exp \left(\hat{a}_{i k}\right)}, \\
    \hat{a}_{i j}=\operatorname{LeakyReLU}(\pi\left(\mathbf{W} {h}_{i}, \mathbf{W} {h}_{j}\right)),
	\label{eq:attention}
\end{gather}
where $\mathcal{N}_i$ is the neighbor set of node $i$, $\hat{a}_{i j}$ denotes the attention score between node $i$ and node $j$ before normalization, $\pi(\cdot)$ represents the correlation function between nodes, $\mathbf{W}$ is the weight matrix, $h_i$ is the feature representation of node $i$, and LeakyReLU is an activation function. The output feature of each node can be calculated as:
\begin{equation}
    h_{i}^{\prime}=\sigma\left(\sum_{j \in \mathcal{N}_{i}} a_{i j} \mathbf{W} h_{j}\right),
\end{equation}
where $\sigma$ denotes the sigmoid activation function.

\subsection{Multimodal Machine Learning}
Multimodal machine learning aims to take full advantage of information from different modalities\cite{poria2017review,liu2021variational}.
Specifically, multimodal machine learning helps to integrate the complementary information across the multimodal data and facilitates extracting similar information to improve model robustness.
Compared with models using single modal data, models using multimodal data always perform better.
How to fuse multimodal information into a unified representation is a primary challenge.
Multimodal machine learning can be realized by multi-kernel learning models~\cite{wen2017multi}, probabilistic graphical models~\cite{zhen2012probabilistic}, neural network models~\cite{wang2020deep}, etc.
Among them, neural networks models have enabled significant performance improvements in many tasks due to their excellent ability to fuse multimodal data.
For instance, Iwana et al.~\cite{iwana2020time} used multimodal CNNs with the local distance-based representation to perform time series classification.
Yang et al.~\cite{yang2020visual} introduced an adaptive weighting algorithm and a multi-head co-attention network to model the association between textual and visual representations in multimodal machine translation.

Recently, multimodal machine learning has been introduced into anomaly detection~\cite{nedelkoski2019anomaly}.
Park et al.~\cite{park2016multimodal} trained a hidden Markov model (HMM) with multimodal data from non-anomalous scenarios and performed anomaly detection for robot manipulation.
Park et al.~\cite{park2018multimodal} leveraged an LSTM-based autoencoder to detect anomalies of an assisted feeding robot.

\section{Methodology}
\label{sec:methodology}
{ This section mainly introduces MST-GAT. MST-GAT uses the multimodal graph attention network and temporal convolution network to capture both temporal and spatial dependencies in multimodal time series.}
\subsection{Problem Definition}
Multimodal time series consist of time series from multiple modalities belonging to the same entity, and each modality can contain one or more time series.
The MTS anomaly detection model is designed to detect anomalies at the timestamp level.
Time series anomaly detection is usually regarded as an unsupervised task, and we assume that there are no anomalies during the training phase.
We formulate this problem as follows.
In the training phase, we train our model in the training set.
The multimodal time series data are composed of $N$ univariate time series with $T$ timestamps, i.e., ${\mathbf{X} = [\mathbf{x}_1, \mathbf{x}_2, \cdots, \mathbf{x}_T]} \in \mathbb{R}^{N \times T}$, and each univariate time series includes $M$ modalities ($1 \le M \le N$).
An observation the multimodal time series at time $t$ ($t \le T$) is denoted by ${\mathbf{x}_t} = [x_{1, t}^{m_1}, x_{2, t}^{m_2}, \cdots, x_{N, t}^{m_N}]^{\top}$, where $x_{i, t}^{m_i}$ ($m_i \in \{1, 2, \cdots, M\}$) denotes the data from the $i$-th univariate time series belonging to the $m_i$-th modality.

In the inference phase, we aim to detect anomalies of in the multimodal time series ${\mathbf{\tilde{X}} = [\mathbf{\tilde{x}}_1, \mathbf{\tilde{x}}_2, \cdots, \mathbf{\tilde{x}}_{T^{\prime}}]} \in \mathbb{R}^{N \times T^{\prime}}$, which is from the same $N$ univariate time series. The length of the multimodal time series is denoted as $T^{\prime}$. The model needs to produce a detection result $\mathbf{\tilde{y}} = [\tilde{y}_1, \tilde{y}_2, \cdots, \tilde{y}_{T^{\prime}}] \in \mathbb{R}^{T^{\prime}}$, where $\tilde{y}_i \in \{0, 1\}$ and $\tilde{y}_i=1$ indicates the $\mathbf{\tilde{x}}_i$ is an anomaly.

\subsection{Overview of MST-GAT}
\begin{figure*}[!t]
\begin{center}
\centerline{\includegraphics [width= 1\textwidth]{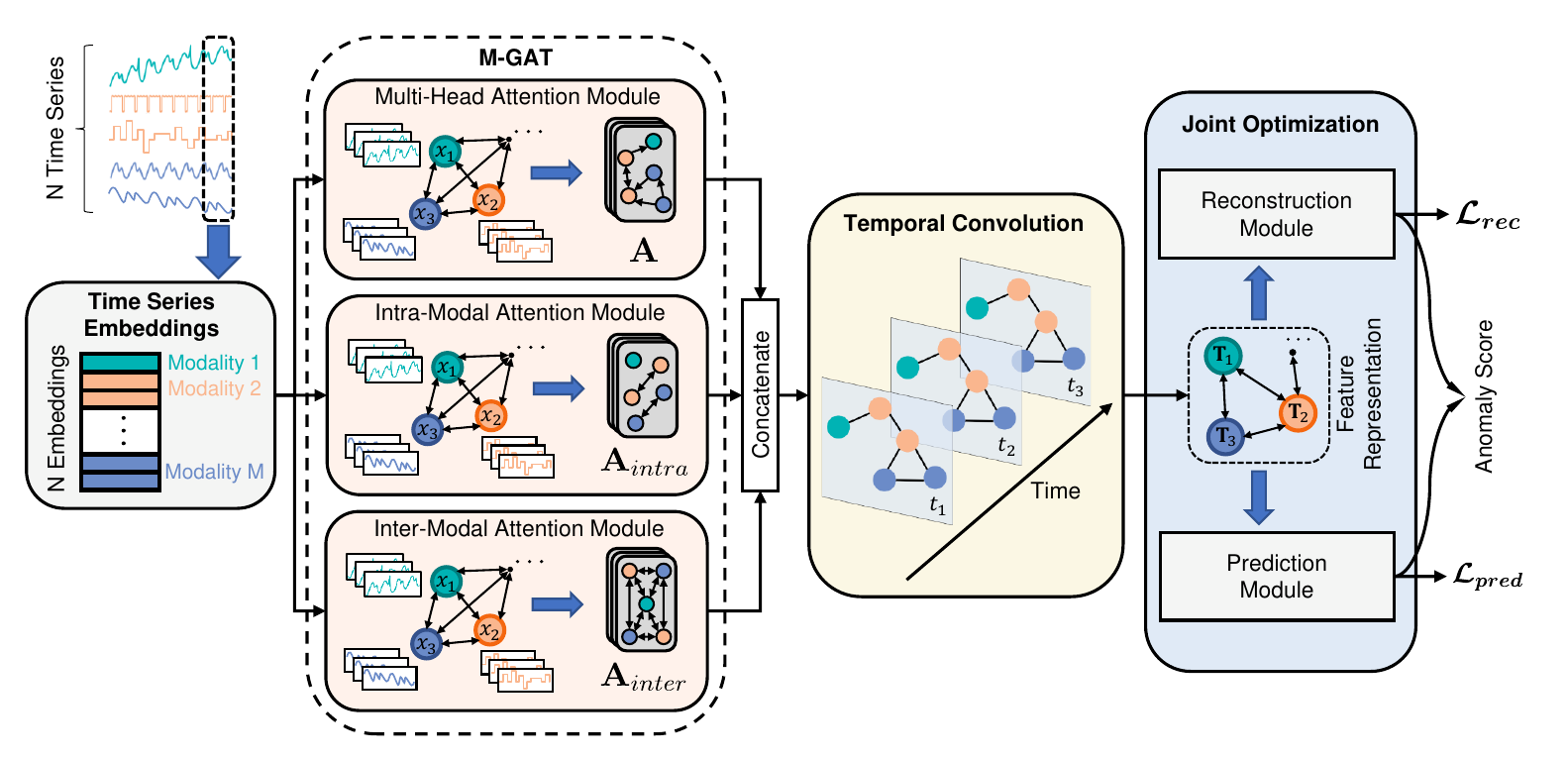}}
\caption{MST-GAT architecture. MST-GAT consists of the M-GAT, a temporal convolution network and a joint optimization network. M-GAT captures spatial and multimodal correlations between different univariate time series. The input data is processed by M-GAT and temporal convolution to explore multimodal correlations and spatial-temporal dependencies. After that, MST-GAT uses a joint optimization network to generate reconstruction probabilities and prediction values. In the training phase, we employ the results of the joint optimization network to optimize MST-GAT.
In the inference phase, the results are further used to calculate anomaly scores for anomaly detection.}
\label{fig:overview}
\end{center}
\end{figure*}
MST-GAT is formulated as a graph structure that treats each sensor as a node and their relationships as edges. It models complex multimodal and spatial-temporal relationships between the entire time series for MTS anomaly detection.
As shown in Figure~\ref{fig:overview}, the architecture of MST-GAT involves four parts:
\begin{itemize}
    \item \textbf{Graph Structure Learning}. It uses time series embeddings (characterizes the inherent properties of each time series) to learn a graph structure in the spatial dimension;
    \item \textbf{Multimodal Graph Attention Network (M-GAT)}. It captures intra- and inter-modal relations explicitly with multi-head attention module and additional relational attention modules (intra- and inter-modal attention);
    \item \textbf{Temporal Convolution Network}. It leverages convolutional structures on the time axis to capture temporal dependencies of time series;
    \item \textbf{Joint Optimization and Anomaly Score}. MST-GAT optimizes both reconstruction and prediction targets, and then identifies anomalies with the anomaly scores. The anomaly scores are further used to interpret detected anomalies.
\end{itemize}

\subsection{Graph Structure Learning}
In multimodal time series, different modalities can exhibit a variety of properties, and they are correlated in complex ways. Therefore, we prefer to use a flexible representation for each time series to capture potential correlations among multiple modalities.
In this paper, we introduce time series embeddings to construct the flexible graph structure for the multi-head attention module.
Consider a graph structure $\mathcal{G}$ with $N$ nodes for multimodal time series, where each node stores the representation of a univariate time series.
The edges between nodes indicate the dependency between different time series.
The set of neighboring nodes of node $i$ are denoted as $\mathcal{N}_i =\{ {j \mid \mathbf{A}_{ij} > 0} \}$.
We define the time series embedding $\mathbf{v}_i \in \mathbb{R}^{d}$ for node $i$ to characterize its inherent properties, where $i \in \{1, 2, \cdots, N\}$ and $d$ is the embedding dimension.
The time series embeddings are further used to construct the adjacency matrix ${\mathbf{A}}$ for the multi-head attention module.
The adjacency matrix can be expressed as:
\begin{gather}
    \mathbf{A}_{ij}=\mathds{1}~~\left\{j \in \operatorname{TopK}\left(\left\{ e_{ik} | k \in \mathcal{C}_{i}\right\}\right)\right\}, \label{eq:adj} \\
    e_{ij} = \operatorname{sim}({\mathbf{v}_i, \mathbf{v}_j})=\frac{\mathbf{v}_i \cdot \mathbf{v}_j}{\left\|\mathbf{v}_i\right\| \times \left\|\mathbf{v}_j\right\|},
\end{gather}
where Eq.~\ref{eq:adj} denotes if $j \in \mathbf{C}_{i}$, then $\mathbf{A}_{ij}=1$, otherwise $\mathbf{A}_{ij}=0$, $\mathcal{C}_i = \{1, 2, \cdots, N\}$ is the candidate set, $\operatorname{sim}(\cdot)$ is the cosine similarity, $e_{ij}$ is the cosine similarity between node $i$ and node $j$, and TopK represents the index of the largest $K$ cosine similarity of node $i$ selected from the candidate set. Specifically, we first compute $e_{ij}$, the cosine similarity between the embedding vectors. Then, we select the top $K$ similar nodes from the candidate set to construct a sparse direct graph, and parameter $K$ controls the sparseness of the graph structure.

\subsection{M-GAT in Spatial Dimension}
For the training time series $\mathbf{X}$, we use the sliding window with length $w$ to produce a fix-length input at each time. We define $\hat{\mathbf{X}}$ as the input of M-GAT at time $t$:
\begin{equation}
    \hat{\mathbf{X}} = [\mathbf{x}_{t-w+1}, \mathbf{x}_{t-w+2}, \cdots, \mathbf{x}_{t}] \in \mathbb{R}^{N \times w},
\end{equation}
and we process the test time series $\mathbf{\tilde{X}}$ in the same way.

\begin{figure}[!t]
\begin{center}
\centerline{\includegraphics [width= 0.6\columnwidth]{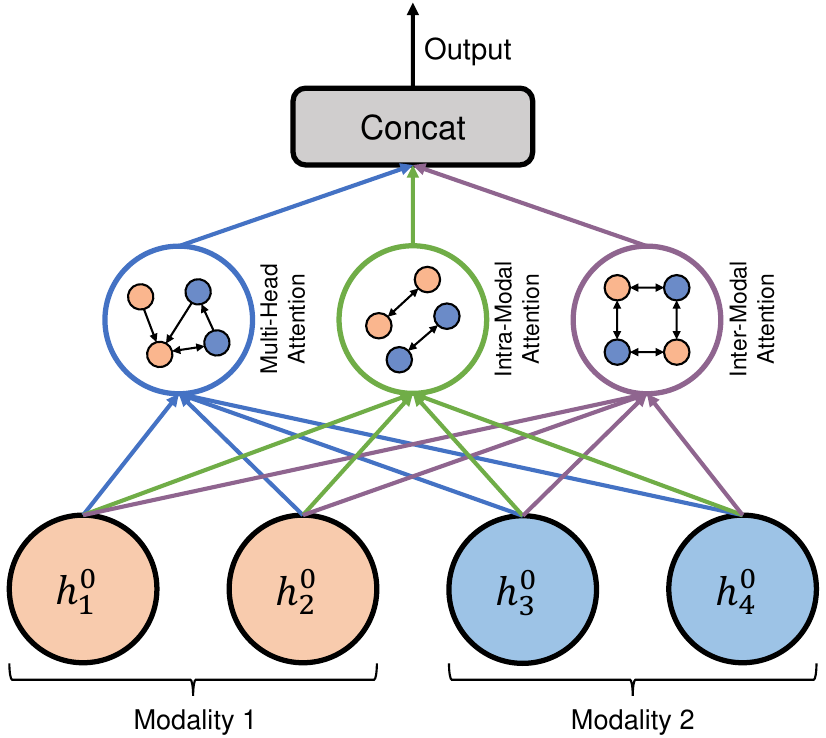}}
\caption{The structure of the M-GAT. It includes three attention modules, i.e., multi-head attention, intra- and inter-modal attention. ${h}_1^0$, ${h}_2^0$, ${h}_3^0$ and ${h}_4^0$ represent the input features of four different univariate time series.
The multi-head attention module models the modality-independent spatial relationships among multimodal time series. The intra- and inter-modal attention modules capture the multimodal correlation between different time series.}
\label{fig::rgat}
\end{center}
\vskip -0.2in
\end{figure}

Suppose $\mathbf{H}^{l}$ denotes the feature representation of M-GAT at layer $l$.
The initial input of M-GAT is $\mathbf{H}^{0}=({\hat{\mathbf{X}} \mathbf{W}_{in})\ ||\ \mathbf{V}}$, where $\mathbf{W}_{in} \in \mathbb{R}^{w \times d}$ is the learnable transformation of input data, and $||$ represents concatenation.
The architecture of the proposed M-GAT is shown in Figure~\ref{fig::rgat}. It consists of three attention modules, i.e., multi-head attention, intra- and inter-modal attention.
The multi-head attention module focuses on modeling the modality-independent spatial relationships among multimodal time series, while intra- and inter-modal attention modules concentrate on capturing the multimodal correlation between different time series.

The multi-head attention module updates the representation of each node by aggregating the representations of its neighbors, which is formulated as:
\begin{equation}
     h_{att_{i}}^{l+1}=||_{s=1}^{S}\sum\limits_{j\in \mathcal{N}_i}\alpha_{ij}^{ls}\mathbf{W}_{att}^{ls}h_j^l,
\label{eq:8}
\end{equation}
\begin{equation}
\alpha_{ij}^{ls}=\operatorname{attention}(i,j),
\label{eq:9}
\end{equation}
where $h_{j}^{l} \in \mathbf{H}^{l}$ is the representation of $j$-th node at layer $l$, $h_{att_{i}}^{l+1}$ is the feature of $i$-th node at layer $l+1$, $S$ denotes the number of attention heads, $||$ represents concatenation, $\alpha_{ij}^{ls}$ represents an attention score calculated by the $s$-th attention head at layer $l$ between node $i$ and node $j$, $\mathbf{W}_{att}^{ls}$ denotes the weight matrix of $s$-th attention head at layer $l$, and $\operatorname{attention}(i,j)$ represents the scaled dot-product attention~\cite{vaswani2017attention}.

Previous methods employ GATs to aggregate the representations of neighboring nodes according to the adjacent matrix~\cite{deng2021graph}.
However, these methods fail to take multimodal dependency of time series into consideration, which may lose some important information on multimodal correlations.
Intuitively, neighboring nodes with different relations should have different influences on the center node.
We extend the multi-head attention module with two additional relational attention modules, i.e., intra- and inter-modal attention, to integrate multimodal time series more effectively.
The adjacency matrix of these two relational attention modules are defined as follows:
\begin{align}
    \mathbf{A}^{ij}_{intra}=\mathds{1} \left\{j \in \mathbf{C}_{intra}^{i} \right\}, \label{eq:adj1} \\
    \mathbf{A}^{ij}_{inter}=\mathds{1} \left\{j \in \mathbf{C}_{inter}^{i} \right\},
\label{eq:10}
\end{align}
where $\mathbf{C}_{intra}^{i}=\left\{j | m_i=m_j\right\}$ and $\mathbf{C}_{inter}^{i}=\left\{j | m_i \neq m_j\right\}$ are the candidate sets.
That is, $\mathbf{C}_{intra}^{i}$ contains nodes that belong to the same modality as node $i$, while $\mathbf{C}_{inter}^{i}$ consists of nodes that belong to different modalities from node $i$.
In general, $\mathbf{C}_{intra}^{i}$ and $\mathbf{C}_{inter}^{i}$ are calculated by Eq.~\ref{eq:adj1} and Eq.~\ref{eq:10}, but if $|\mathbf{C}_{intra}^{i}|>K$, the index of the largest $K$ cosine similarity values will be selected by TopK operation. Similarly, if $|\mathbf{C}_{inter}^{i}|>K$, the TopK operation will also be applied to $\mathbf{C}_{inter}^{i}$:
\begin{align}
\mathbf{C}_{intra}^{i}=\left\{\operatorname{TopK}\left(\left\{e_{ik}|k \in \mathbf{C}_{intra}^{i} \right\}\right)\right\}, \\
\mathbf{C}_{inter}^{i}=\left\{\operatorname{TopK}\left(\left\{e_{ik}|k \in \mathbf{C}_{inter}^{i} \right\}\right)\right\}.
\end{align}

Then, we use these two relational attention modules to model the multimodal dependency among time series explicitly. We compute the feature of the intra-modal attention module as:
\begin{gather}
  h^{l+1}_{intra_{i}}=\sum\limits_{j \in \mathcal{N}_{intra_{i}}}\beta^{lj}_{intra_{i}} \mathbf{W}_{intra}^l h_j^l, \\
  \beta^{lj}_{intra_{i}}=\frac{\operatorname{exp}(g^{lj}_{intra_{i}})}{\sum\limits_{k \in {\mathcal{N}_{intra_{i}}}}\operatorname{exp}(g^{lk}_{intra_{i}})}, \\
  g^{lj}_{intra_{i}}=\sigma(\operatorname{ReLU}((\mathbf{V}_{i}  || \mathbf{V}_{j})  \mathbf{W}_{intra1}^{l}+b_{intra1}^{l}) \mathbf{W}_{intra2}^{l}),
\label{eq:14}
\end{gather}
where $h_{intra_{i}}^{l+1}$ is the feature of $i$-th node at layer $l+1$,
$\mathcal{N}_{intra_i} =\{{j \mid \mathbf{A}_{intra}^{ij} > 0}\}$ denotes the intra-modal neighbor set of node $i$,
$\beta^{lj}_{intra_{i}}$ represents the attention score at layer $l$ between node $i$ and node $j$,
$\mathbf{W}_{intra}^l$, $\mathbf{W}_{intra1}^l$ and $\mathbf{W}_{intra2}^l$ are weight matrixes at layer $l$,
and $b_{intra1}^{l}$ is the bias vectors at layer $l$.
The computation of $h^{l+1}_{inter_i}$ resembles the way of computing $h^{l+1}_{intra_i}$, and $\mathcal{N}_{inter_i} =\{{j \mid \mathbf{A}_{inter}^{ij} > 0}\}$ is the inter-modal neighbor set of node $i$.
We incorporate the $h_{att_i}^{l+1}$, $h_{intra_i}^{l+1}$ and $h_{inter_i}^{l+1}$ into the final representation $h_i^{l+1}$:
\begin{gather}
 h_i^{l+1}=\operatorname{ReLU}(\mathbf{W}_{out}^{l+1}o_{i}^{l+1}+b_{out}^{l+1}),  \label{eq:16-1} \\
 o_{i}^{l+1}=h_{att_i}^{l+1}\ ||\ h_{intra_i}^{l+1}\ ||\ h_{inter_i}^{l+1},
 \label{eq:16-2}
\end{gather}
where $h_i^{l+1}$ is the final representation of node $i$ at layer $l+1$, $\mathbf{W}_{out}^{l+1}$ is the weight matrix at layer $l+1$, $b_{out}^{l+1}$ is the bias vectors at layer $l+1$, $||$ denotes concatenation, and $o_{i}^{l+1}$ is the intermediate feature at layer $l+1$ by concatenating $h_{att_i}^{l+1}$, $h_{intra_i}^{l+1}$ and $h_{inter_i}^{l+1}$.

\subsection{Convolution in Temporal Dimension}
The multimodal graph attention captures the neighbors' information of each node in the spatial dimension, while the temporal convolution network applies the stand convolution on the time dimension to capture the temporal dynamic.
The input of the temporal convolution network is the graph-level representation $\mathbf{H}^{L_{gat}}$, where $L_{gat}$ is the number of layers in M-GAT. The temporal-level representation is calculated as:
\begin{equation}
{\mathbf{T}}^{l+1}=\operatorname{ReLU}\left(\Phi *\left(\operatorname{ReLU}\left({\mathbf{T}}^{l}\right)\right)\right),
\label{eq:17}
\end{equation}
where $\mathbf{T}^{l+1}$ denotes the temporal-level representation at layer $l+1$, $*$ is the standard convolution operation, $\Phi$ is the kernel size, and ReLU is an activation function. The temporal convolution network updates the features of nodes by incorporating information from adjacent time slices, so it can well capture the temporal dynamics.

\subsection{Joint Optimization and Anomaly Score}
The input of the reconstruction and prediction modules is the output of the temporal convolution network.
For clarity, we set $\mathbf{\mathcal{X}}_{t}\!=\!\mathbf{T}^{L_{tem}}$ as the input of reconstruction and prediction modules at time $t$, where $L_{tem}$ denotes the number of layers in temporal convolution.
MST-GAT combines the advantages of reconstruction and prediction modules.
The reconstruction module captures the data distribution of the whole time series, and the prediction module forecasts the observations at the next timestamp.
We optimize MST-GAT with two tasks, i.e., reconstruction and prediction tasks. The loss function contains two optimization objectives, which are defined as:
\begin{equation}
\mathcal{L} = \gamma_{1}\mathcal{L}_{rec}+(1-\gamma_{1})\times \mathcal{L}_{pred},
\label{eq:18}
\end{equation}
where $\mathcal{L}_{rec}$ represents the loss function for the reconstruction module, $\mathcal{L}_{pred}$ denotes the loss function for the prediction module, and $\gamma_{1}$ is a hyperparameter that balances the reconstruction and prediction modules.

\subsubsection{Reconstruction Module}
The goal of the reconstruction module is to learn the reconstruction probability of the input data. Inspired by OmniAnomaly~\cite{su2019robust}, we use variational autoencoder (VAE) to reconstruct $\mathcal{G}_{t}$. Given the input $\mathbf{\mathcal{X}}_{t}$, VAE uses the conditional distribution $p_\psi(\mathbf{\mathcal{X}}_{t}|{z}_{t})$ to reconstruct $\mathcal {X}_{t}$, where $z$ is the latent representation.
The goal of training the reconstruction module is to maximize the posterior distribution of ${z}_{t}$:
\begin{equation}
p_\psi({z}_{t}|\mathbf{\mathcal{X}}_{t})=p_\psi(\mathbf{\mathcal{X}}_{t}|{z}_{t})p_{\psi}({z}_{t})/p_{\psi}(\mathbf{\mathcal{X}}_{t}),
\label{eq:19}
\end{equation}
where $p_{\psi}(\mathbf{\mathcal{X}}_{t})$ is the reconstruction probability of $\mathbf{\mathcal{X}}_{t}$. Let $p_{\psi}(\mathbf{\mathcal{X}}_{t})$ = $\{ p_i | i = 1, 2, \cdots, N \}$, where $p_i$ denotes the reconstruction probability of $i$-th univariate time series. At each timestamp, the combined model will produce two inference results. The reconstruction probability $p_{\psi}(\mathbf{\mathcal{X}}_{t})$ can be defined as follows:
\begin{equation}
p_\psi(\mathbf{\mathcal{X}}_{t})=\int{p_\psi({z}_{t})p_\psi(\mathbf{\mathcal{X}}_{t}|{z}_{t})d{z}_{t}}.
\label{eq:20}
\end{equation}
The above equation is difficult to calculate, and we need a new model $q_\rho({z}_{t}|\mathbf{\mathcal{X}}_{t})$ to approximate the $p_\psi({z}_{t}|\mathbf{\mathcal{X}}_{t})$. Given the encoder model $q_\rho({z}_{t}|\mathbf{\mathcal{X}}_{t})$ and the decoder model $p_\psi(\hat{\mathcal {X}_{t} }|{z}_{t})$, the reconstruction loss is formulated as:
\begin{equation}
\begin{aligned}
\mathcal{L}_{rec} = -\mathrm{E}_{q_{\rho}({z}_{t}|\mathbf{\mathcal{X}}_{t})}[log{p_{\psi}(\mathbf{\mathcal{X}}_{t}|{z}_{t})}] \\
+ D_{KL}(q_{\rho}({z}_{t}|\mathbf{\mathcal{X}}_{t})||p_{\psi}({z}_{t})),
\end{aligned}
\end{equation}
where $\mathrm{E}_{q_{\rho}({z}_{t}|\mathbf{\mathcal{X}}_{t})}[log{p_{\psi}(\mathbf{\mathcal{X}}_{t}|{z}_{t})}]$ denotes the log-likelihood expectation of $\mathbf{\mathcal{X}}_{t}$. $D_{KL}$ represents the KL divergence. Negative $\mathcal{L}_{rec}$ is an estimation of the lower bound of $\log{p_\psi(\mathbf{\mathcal{X}}_{t})}$.

\subsubsection{Prediction Module}
The prediction module uses $\mathbf{\mathcal{X}}_{t}$ to predict the observations of the next timestamp. We use a multi-layer perceptron (MLP) network as the prediction module behind the temporal convolution network. The prediction loss can be defined as:
\begin{align}
& \mathcal{L}_{pred}=\frac{1}{T-w}\sqrt{\sum_{i=1}^{N}({x}_{i,t+1}-\hat{x}_{i,t+1})^2},
\label{eq:loss}
\end{align}
where ${x}_{i, t+1}$ denotes the ground truth value of $i$-th time series at time $t+1$, and $\hat{x}_{i,t+1}$ is the forecast of $i$-th time series at time $t+1$.

\subsubsection{Anomaly Score and Inference}
\label{sec:341}
At each timestamp, the reconstruction module and the prediction module generate the reconstruction probability $p_i$ and the forecast $\mathbf{\hat{x}}_i$, respectively, where $\mathbf{\hat{x}}_{i}$ denotes the prediction value of $i$-th univariate time series.
The anomaly score of MST-GAT balances the weights of these two modules.
The final anomaly score of each timestamp is the sum of anomaly scores for each time series. Specifically, the anomaly score is formulated as:
\begin{align}
\operatorname{score} = \sum_{i=1}^N{\frac{(1 - p_{i}) + \gamma_{2} \times (\mathbf{x}_{i} - \mathbf{\hat{x}}_{i})^2}{1 + \gamma_{2}}},
\label{eq:score}
\end{align}
where $(\mathbf{x}_{i} - \mathbf{\hat{x}}_{i})^2$ is the square error between the forecast $\mathbf{\hat{x}}_i$ and the ground truth $\mathbf{x}_i$, and $\gamma_{2}$ is the hyperparameter introduced to balance the two modules, which is selected by the validation set.

In the inference phase, the detection rule is that if the anomaly score at a timestamp is greater than the defined anomaly threshold, this timestamp will be marked as ``abnormal'', otherwise ``normal''. We adopt the peaks-over-threshold (POT) algorithm~\cite{siffer2017anomaly} to select the anomaly threshold over the validation set. Finally, the overall training and inference process of MST-GAT is summarized in Algorithm~\ref{algorithm1}.

\renewcommand{\algorithmicrequire}{\textbf{Input:}}
\renewcommand{\algorithmicensure}{\textbf{Output:}}
\begin{algorithm}[!t]
    \footnotesize
    \caption{Training and Inference Procedures of MST-GAT}
    \centerline{\textbf{Training Procedure}}
    \begin{algorithmic}[1]
    \REQUIRE  Multimodal training time series ${\mathbf{X} = [\mathbf{x}_1, \mathbf{x}_2, \cdots, \mathbf{x}_T]}$, training epochs $I$, batch size $M$ and hyperparameters $\gamma_{1}, \gamma_{2}$.
    \STATE Randomly initialize parameter $W_{model}$ ($W_{model}$ includes all learnable parameters in MST-GAT);
    \FOR{epoch $i \in 1, 2, \dots, I$}
        \STATE Calculate $\mathbf{H}^{L_{gat}}$ in spatial dimension by M-GAT; ~~// Eq.~\ref{eq:16-1}
        \STATE Calculate $\mathbf{T}^{L_{tem}}$ in temporal dimension by temporal convolution; ~~// Eq.~\ref{eq:17}
        \STATE Calculate the reconstruction probability via reconstruction module; ~~// Eq.~\ref{eq:20}
        \STATE Calculate the prediction  value via prediction module;
        \STATE Minimize the joint loss function to optimize $W_{model}$; ~~// Eq.~\ref{eq:18}
    \ENDFOR
    \RETURN The optimized model parameter $W_{model}$.
    \end{algorithmic}

    \centerline{\textbf{Inference Procedure}}
    \begin{algorithmic}[1]
    \REQUIRE  Multimodal testing time series ${\mathbf{\tilde{X}} = [\mathbf{\tilde{x}}_1, \mathbf{\tilde{x}}_2, \cdots, \mathbf{\tilde{x}}_{T^{\prime}}]}$, model parameter $W_{model}$ and hyperparameters $\gamma_{1}$, $\gamma_{2}$.
    \FOR {\textbf{each} $\mathbf{\tilde{x}}_i$ }
    \STATE Calculate the anomaly score of $\mathbf{\tilde{x}}_i$; ~~// Eq.~\ref{eq:score}
    \IF {anomaly score $>$ threshold}
    \STATE $\mathbf{\tilde{x}}_i =$ ``an anomaly'';
    \ELSE
    \STATE $\mathbf{\tilde{x}}_i =$ ``a normal point'';
    \ENDIF
    \ENDFOR
    \RETURN Predicted label list of $\mathbf{\tilde{X}}$.
    \end{algorithmic}
    \label{algorithm1}
\end{algorithm}

\section{Experiments}\label{sec:exp}
{
In this section, we conduct comprehensive experiments to demonstrate the effectiveness of MST-GAT. We first introduce the four commonly-used public datasets. Next, we evaluate MST-GAT on these datasets and show that MST-GAT performs better or on par with a range of baselines and substantially outperforms current anomaly detection methods. Then, we perform ablation studies on the key components of the proposed model. Finally, we provide the interpretability of MST-GAT through a case study.
}

\begin{table}[!t]
    \centering
    \caption{Statistics of the four datasets used in the experiments.}
    \begin{tabular}{ ccccccc }
     \toprule
        {Datasets} & {Features} & {Modalities} & {Train} & {Test} & {Anomalies (\%)}\\
     \midrule
        MSL  & 27  & 8   & 58317   & 73729  & 10.72 \\
        SMAP & 55  & 12  & 135183  & 427617 & 13.13 \\
        SWaT & 51  & 8 & 496800  & 449919 & 11.98 \\
        WADI & 123 & 8  & 1048571 & 172801 & 5.99  \\
     \bottomrule&
    \end{tabular}
    \label{table:dataset}
\end{table}

\subsection{Datasets}
This experiment involves four benchmark datasets. We show the statistics in Table~\ref{table:dataset} and briefly introduce them in the following.

Mars Science Laboratory rover (MSL)~\cite{hundman2018detecting} and
Soil Moisture Active Passive satellite (SMAP)~\cite{hundman2018detecting} are real-world datasets acquired from the spacecraft. These datasets are annotated by experts of NASA. Each dataset includes pre-segmented training and test sets. The training set is collected from the normal data, and the test set includes labeled anomalies.
Secure Water Treatment (SWaT)~\cite{goh2016dataset} dataset is collected from the scaled-down water treatment testbed with 51 sensors, consisting of seven days of normal operation and four days of simulated attack scenarios. These simulated attacks include different durations and diverse attack targets.
Water Distribution (WADI)~\cite{dataset_wadi} is a dataset acquired from a reduced water distribution testbed comprising 123 sensors. It includes two weeks under normal operation as a training set and two days with attack scenarios as a test set.

\begin{figure}[!t]
    \begin{center}
    \centerline{\includegraphics [width= 0.6\columnwidth]{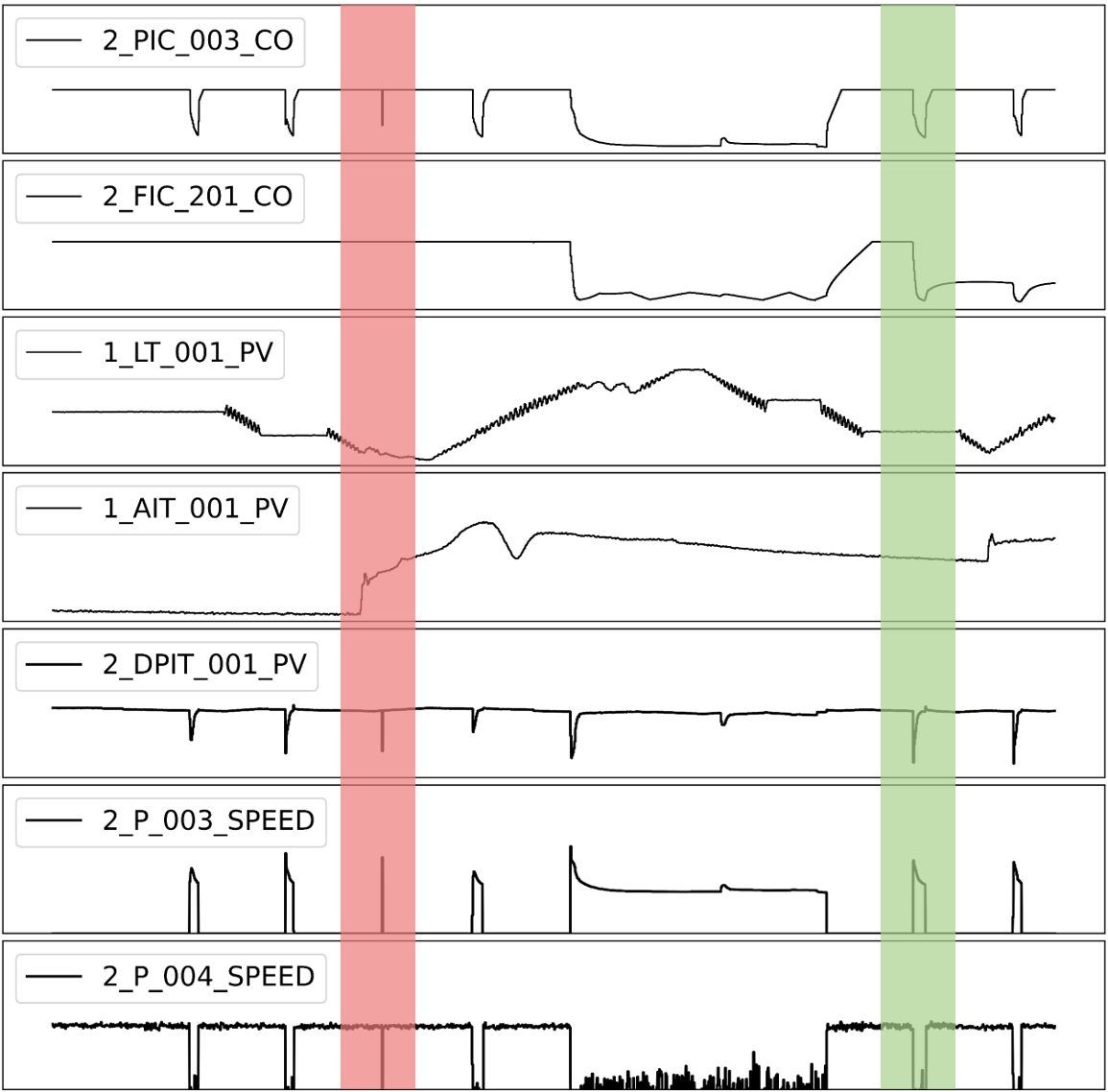}}
    \caption{An example of multimodal time series data. The red shaded area (left) indicates anomalies, and the green shaded area (right) indicates normal values. Time series with the same suffix belongs to the same modality (e.g., $2\_P\_003\_SPEED$ and $2\_P\_004\_SPEED$ belong to the speed modality).}
    \label{fig::dataset_introduction}
    \end{center}
\end{figure}

Figure~\ref{fig::dataset_introduction} shows an example of multimodal time series on WADI dataset.
In the green shaded area (right), all sensor values have obvious fluctuations except $2\_FIC\_201\_CO$ and $1\_LT\_001\_PV$, but the system is still in a normal state as these time series maintain a consistent trend.
Nevertheless, in the red shaded area (left) segment, sensor $1\_AIT\_001\_PV$ behaves an inconsistent pattern compared with other univariate time series, indicating a potential problem in this sensor.

\subsection{Baselines}
{
We compare MST-GAT with eight popular MTS anomaly detection methods. They can be divided into two groups: 1) four monomodal methods, including PCA~\cite{pca}, AE~\cite{provotar2019unsupervised}, DAGMM~\cite{dagmm}, and LSTM-VAE~\cite{park2018multimodal}; 2) four multimodal methods, including MAD-GAN~\cite{li2019mad}, OmniAnomaly~\cite{su2019robust}, USAD~\cite{audibert2020usad}, and GDN~\cite{deng2021graph}. The details are presented as follows.
}
\begin{itemize}
    \item \textbf{\textsc{PCA:}} Principal component analysis projects the high-dimensional representation to the low-dimensional representation, and the reconstruction error of the projection is used for computing the anomaly score.
    \item \textbf{\textsc{AE:}} Autoencoder includes an encoder and a decoder and uses the reconstruction error to detect anomalies. The encoder compresses the input data into a hidden vector, and the decoder uses the vector to reconstruct the input data.
    \item \textbf{\textsc{DAGMM:}} Deep autoencoding Gaussian model combines the deep autoencoder and the Gaussian mixture model to generate the low-dimensional feature. DAGMM is a classic reconstruction-based method, which employs the reconstruction error as the anomaly score.
    \item \textbf{\textsc{LSTM-VAE:}} LSTM-VAE substitutes the fully connected network in variational autoencoder with LSTM, which can better capture the temporal dependence.
    \item \textbf{\textsc{MAD-GAN:}} Multivariate anomaly detection strategy with GAN leverages LSTM-RNN as the generator and discriminator for time series anomaly detection.
    \item \textbf{\textsc{OmniAnomaly:}} OmniAnomaly adopts a stochastic recurrent neural network for time series anomaly detection and employs the reconstruction probability to explain the detected anomalies.
    \item \textbf{\textsc{USAD:}} Unsupervised anomaly detection is an autoencoder-based framework and is trained in an adversarial fashion. The autoencoder in USAD makes the adversarial training more stable.
    \item \textbf{\textsc{GDN:}} Graph deviation network is an unsupervised anomaly detection method and uses the graph attention mechanisms to perform structure learning in multivariate time series and interprets the detected anomaly by attention weights.
\end{itemize}

\subsection{Evaluation Metrics for MTS Anomaly Detection}
We use precision ($\text{Prec}$), recall ($\text{Rec}$), F1-score ($\text{F1}$), and the area under the ROC curve (AUC) as the evaluation metrics. The ROC curve represents a plot of true positive rate between false positive rate, and AUC is defined as the area under the ROC curve. Precision is the ratio of correctly detected anomalies to all detected anomalies.
Recall is the ratio of correctly detected anomalies to all actual anomalies. F1-score can comprehensively consider the value of precision and recall. For clarity, we denote true positives, false positives, and false negatives as TP, FP, and FN, respectively. Precision, recall and F1-score are formulated as:
\begin{gather}
\text{Prec}=\frac{\text{TP}}{\text{TP}+\text{FP}},\\
\text{Rec}=\frac{\text{TP}}{\text{TP} +\text{FN}},\\
\text{F1}=\frac{2 \times \text{Prec} \times \text{Rec}}{\text{Prec} + \text{Rec}}.
\end{gather}

\begin{table}[!t]
    \centering
    \caption{Comparison with existing methods in terms of precision (\%), recall (\%) and F1-score (\%) on four datasets. The best results are highlighted in bold, and the second-best results are marked underlined.
    { The up arrow ($\uparrow$) indicates a significant improvement of our method compared to the baseline.}}
    \label{tab:performance}
    \renewcommand\arraystretch{1}
    \setlength{\tabcolsep}{1.2mm}{
    \scriptsize
    \begin{tabular}{@{}lccccccccccccc@{}}
    \toprule
                     & \multicolumn{3}{c}{\textbf{MSL}}              & \multicolumn{3}{c}{\textbf{SMAP}}             & \multicolumn{3}{c}{\textbf{SWaT}}             & \multicolumn{3}{c}{\textbf{WADI}}           & \multicolumn{1}{c}{}                                                                           \\ \cmidrule(r){2-4} \cmidrule(r){5-7} \cmidrule(r){8-10} \cmidrule(r){11-13}
    \textbf{Method}             & {Prec} & {Rec} & {F1}    & {Prec} & {Rec} & {F1}    & {Prec} & {Rec} & {F1}    & {Prec} & {Rec} & {F1}                                                                           & \multicolumn{1}{c}{\multirow{-2}{*}{\begin{tabular}[c]{@{}c@{}}Wilcoxon\\ test\end{tabular}}}  \\ \midrule
    PCA              & 29.37         & 24.14        & 26.50          & 28.84         & 19.93        & 23.57          & 24.92         & 21.63        & 23.16          & 39.53         & 5.63         & 9.86         & $\uparrow$                                                                                     \\
    AE               & 71.66         & 50.08        & 58.96          & 72.16         & 79.95        & 75.86          & 72.63         & 52.63        & 61.03          & 34.35         & 34.35        & 34.35        & $\uparrow$                                                                                     \\
    DAGMM            & 49.11         & 55.62        & 52.16          & 58.45         & 90.58        & 71.05          & 27.46         & 69.52        & 39.37          & 54.44         & 26.99        & 36.09        & $\uparrow$                                                                                     \\
    LSTM-VAE         & 52.57         & 95.46        & 67.80          & 85.51         & 63.66        & 72.98          & 96.24         & 59.91        & 73.85          & 87.79         & 14.45        & 24.82        & $\uparrow$                                                                                     \\   
    MAD-GAN          & 85.17         & 89.91        & 87.48          & 80.49         & 82.14        & 81.31          & 98.97         & 63.74        & 77.54          & 41.44         & 33.92        & 37.30        & $\uparrow$                                                                                     \\
    OmniAno.         & 88.67         & 91.17        & 89.90          & 74.16         & 97.76        & 84.34          & 98.25         & 64.97        & 78.22          & 99.47         & 12.98        & 22.96        & $\uparrow$                                                                                     \\
    USAD             & 93.08         & 89.17        & {\ul 91.08}    & 90.96         & 85.29        & 88.03          & 98.51         & 66.18        & 79.17          & 64.51         & 32.20        & 42.96        & $\uparrow$                                                                                     \\
    GDN              & 91.35         & 86.12        & 88.66          & 89.32         & 88.72        & {\ul 89.02}    & 99.35         & 68.12        & {\ul 80.82}    & 97.50         & 40.19        & {\ul 56.92}  & $\uparrow$                                                                                     \\ \midrule
    MST-GAT & 95.06         & 89.10        & \textbf{91.98} & 91.26         & 89.83        & \textbf{90.54} & 98.73         & 72.41        & \textbf{83.55} & 98.24         & 43.51        & \textbf{60.31}        & /  \\ \bottomrule
    \end{tabular}}
\end{table}

\subsection{Experimental Setup}
Our methods are implemented with PyTorch and trained in a Ubuntu server with Intel(R) Xeon(R) CPU E5-2640 @ 2.50GHz and an NVIDIA 2080ti GPU.
We use the Adam optimizer for training, and the learning rate is set to $ 1 \times 10^{-3}$.
The whole network is trained with a batch size of $32$ and a total of $60$ epochs.
The embedding dimension $d$ is set to $128$ for all datasets.
We empirically set the sliding window size to $32$, the kernel size of temporal convolution to $16$, and the number of attention heads $S$ to $4$ for each dataset.
We set the $K$ to $15$, $30$, $30$, and $30$ for MSL, SMAP, SWaT, and WADI, respectively.
The model hyperparameters $\gamma_{1}$ and $\gamma_{2}$ are selected as $0.5$ and $0.8$ through grid search.
We utilize the POT algorithm~\cite{siffer2017anomaly} to set the anomaly threshold over the validation dataset.
In the inference phase, any timestamp whose anomaly score exceeds the threshold will be considered as ``abnormal.''

\subsection{Results and Analysis}
\label{ssec:results}
Table~\ref{tab:performance} summarises the comparison of MST-GAT and baselines in terms of accuracy, recall and F1-score on four datasets.
The results show that MST-GAT consistently outperforms the existing baselines on four benchmarks in terms of F1-score.
We can observe that most baselines perform better on MSL and SMAP datasets because they have relatively simpler anomaly patterns and spatial-temporal dynamics, and MST-GAT still outperforms the best baseline by $0.9$ and $1.52$ in terms of F1-score (\%) on MSL and SMAP datasets, respectively.
Besides, most baselines show inferior results on SWaT and WADI datasets, which contain more complex anomalies, but MST-GAT significantly exceeds them by at least $2.73$ in terms of F1-score (\%).
The structure with multimodal graph attention network and temporal convolution network efficiently integrates the information of multimodal time series, enabling MST-GAT to capture the complex spatial-temporal dynamics in multimodal time series.
Not surprisingly, traditional methods (e.g., PCA and DAGMM) do not perform well compared with most deep learning methods, as they are difficult to encode comprehensive information in the time series and lack adequate consideration of the spatial and temporal dependence.
Moreover, the recent USAD and GDN achieve better performance than other baselines. However, GDN is not good at obtaining temporal features from time series. USAD ignores the spatial correlation in the multimodal time series, and problems may occur when the underlying spatial dependence is complex. MST-GAT outperforms GDN that also adopts graph attention networks, showing the feasibility of using the multimodal graph attention network and temporal convolutional network.
{
We employ the Wilcoxon signed-rank test at the 95\% confidence level to identify whether the difference in performance between MST-GAT and other baselines is significant on four datasets. It is observed that our method has a significant performance improvement compared to any of the baseline methods.
}

\begin{figure}[!t]
    \begin{center}
    \centerline{\includegraphics [width= 1\columnwidth]{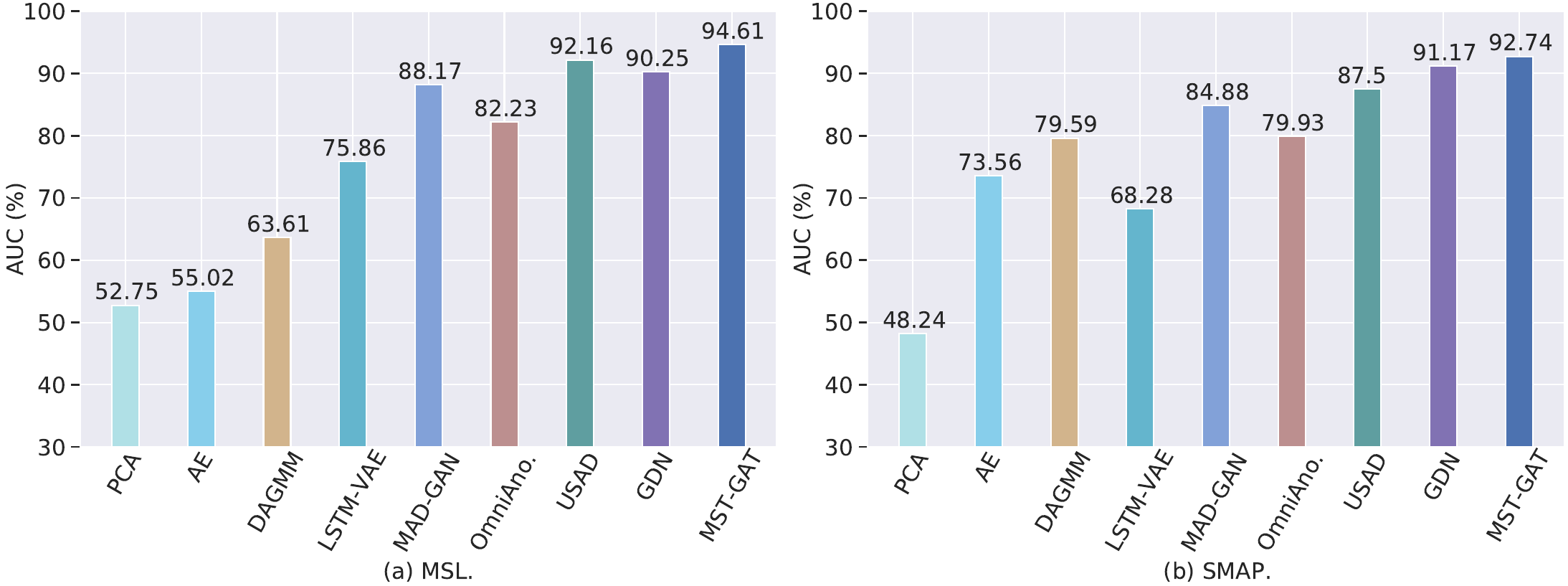}}
    \caption{AUC (\%) results on MSL and SMAP datasets. The larger the better.}
    \label{fig::auc}
    \end{center}
\end{figure}

\begin{figure}[!t]
    \begin{center}
    \centerline{\includegraphics [width= 0.65\columnwidth]{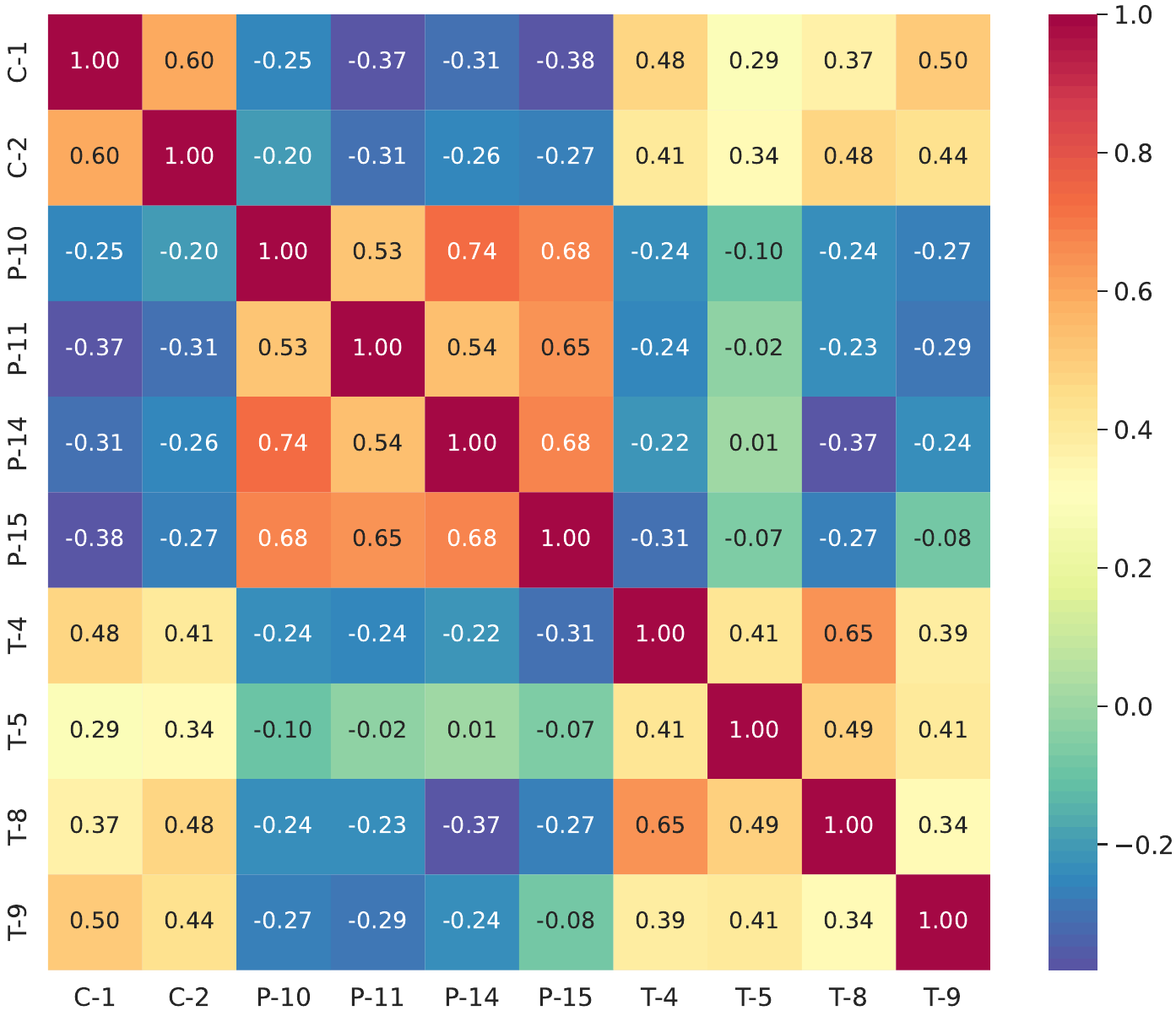}}
    \caption{Illustration of the cosine similarity of time series embeddings between three modalities that are randomly selected from MSL. The same prefix means the same modality (e.g., T-$4$ and T-$5$ belong to the temperature modality). }
    \label{fig::confusion}
    \end{center}
\end{figure}

\begin{figure*}[!t]
\begin{center}
\centerline{\includegraphics [width= 0.93\textwidth]{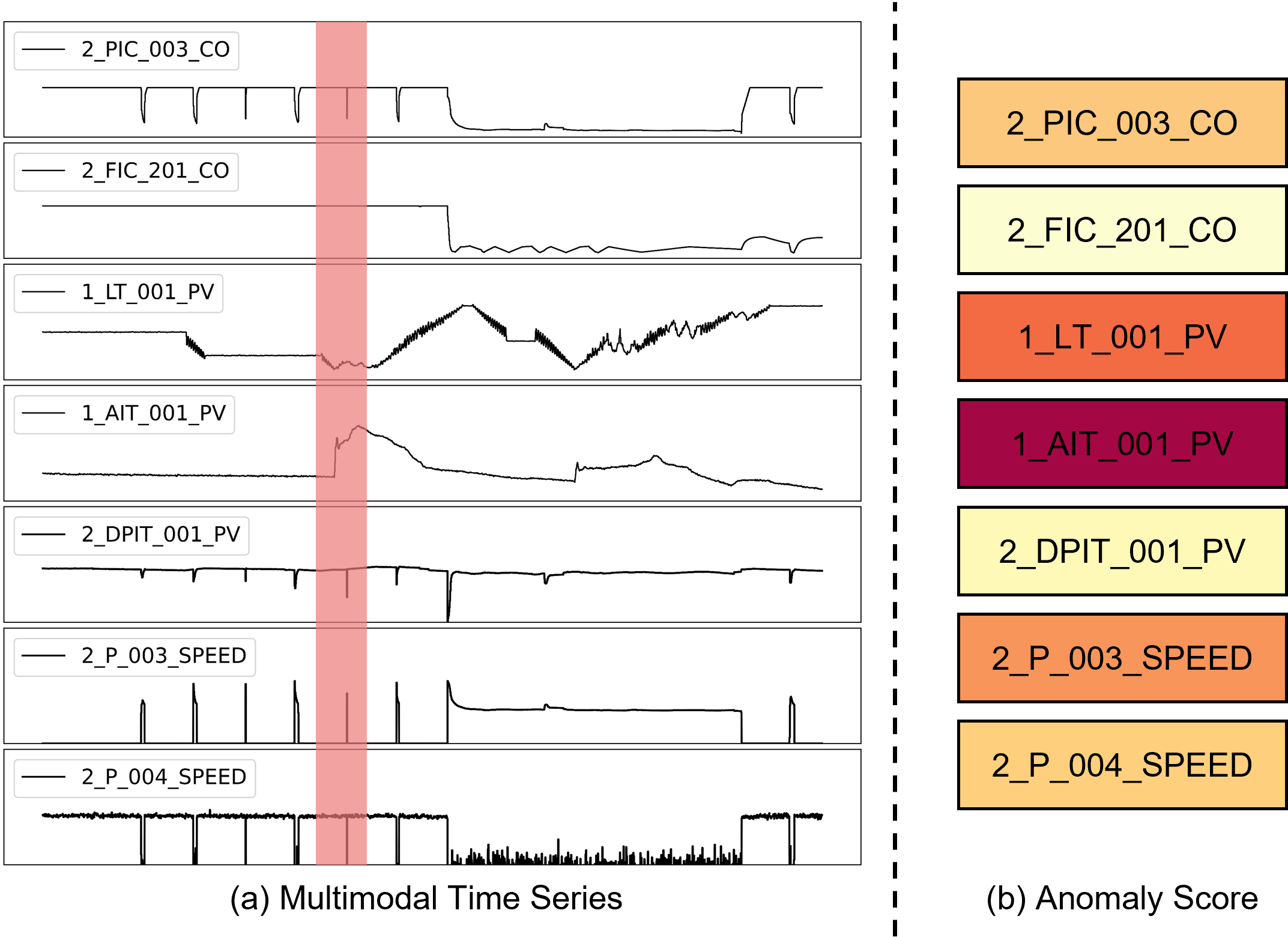}}
\caption{
The analysis of the proposed interpret method on anomalies (red shaded area) of WADI dataset. Time series with the same suffix belongs to the same modality. The darker color indicates the higher anomaly score in the red shaded area. The time series $1\_AIT\_001\_PV$ obtains the highest anomaly score as it shows significant inconsistencies with other time series. Therefore, $1\_AIT\_001\_PV$ is selected by MST-GAT as the sensor most likely to cause the anomaly.
}

\label{fig:interpret}
\end{center}
\end{figure*}

We further calculate the AUC as the performance indicator on MSL and SMAP datasets, as demonstrated in Figure~\ref{fig::auc}.
The proposed MST-GAT consistently outperforms other strong baselines.
We attribute the performance advantage to the effective use of spatial and temporal information in multimodal time series.
By using M-GAT and the temporal convolution network, MST-GAT considers the interaction of intra-modal, inter-modal and temporal information to capture the spatial-temporal correlations among multimodal time series.
The results suggest that the use of multimodal graph attention network and temporal convolution network facilitates MST-GAT to achieve higher true-negative rates and lower false-positive rates in anomaly detection.

The confusion matrix of time series embeddings on MSL dataset is shown in Figure~\ref{fig::confusion}. It is seen that univariate time series belonging to the same modality perform relatively high similarity, which demonstrates the intra-modal consistency on MSL dataset. As different modalities may exhibit different degrees of correlations, modality C and modality T are higher correlated with similarities in the range of $[0.29,0.50]$, while modality P and modality T exhibit lower correlations with absolute similarities in the range of $[0.01,0.37]$. Overall, time series embeddings strongly reflect the intra- and inter-modal correlations, revealing the feasibility of using M-GAT to model the modal dependencies between different time series.

We also conduct experiments to show the interpretability of MST-GAT. Figure~\ref{fig:interpret}(a) demonstrates an anomaly example of the multimodal time series on WADI dataset.
The shaded area is an anomaly interval in the multimodal time series.
MST-GAT gives an average anomaly score of each sensor in the anomaly interval, and the sensor with the highest average anomaly score in the red shaded area is located as the sensor most likely to cause this anomaly.
As depicted in Figure~\ref{fig:interpret}(b), the darker color indicates the higher anomaly score, and $1\_AIT\_001\_PV$ is selected as the sensor that is most likely to cause this anomaly, as it shows an inconsistent trend with other time series.
MST-GAT provides interpretable results that are consistent with human intuition.
The ability of  MST-GAT to interpret anomalies is largely attributed to the multimodal graph attention of MST-GAT, which correctly captures the correlation between features.

\begin{table}[!t]
    \caption{Hyperparameter analysis with F1-score (\%) on SWaT dataset. The best result is marked in bold.}
    \label{tab::sensitive}
    \centering
    \footnotesize
    \renewcommand\arraystretch{1}
    \setlength{\tabcolsep}{5mm}{
    \footnotesize
    \begin{tabular}{ l c c c c}
      \toprule   \diagbox{$\gamma_{1}$}{$\gamma_{2}$}  &   0.4   & 0.6   & 0.8            &   1   \\
      \midrule                0.2             &   82.07 & 82.84 & 83.26          & 82.84 \\
                              0.5             &   82.03 & 83.16 & \textbf{83.55} & 83.45 \\
                              0.8             &   81.97 & 82.50 & 83.35          & 83.01 \\
      \bottomrule
    \end{tabular}}
\end{table}

Furthermore, we perform sensitivity analysis of the hyperparameters on SWaT dataset. We focus on two important hyperparameters $\gamma_{1}$ and $\gamma_{2}$, which are used in the proposed loss function and anomaly score. Table~\ref{tab::sensitive} reports the F1-score (\%) for all combinations of $\gamma_{1}$ from $0.2$ to $0.8$ with increment $0.3$, and $\gamma_{2}$ from $0.4$ to $1$ with increment $0.2$. The results demonstrate that MST-GAT is insensitive to $\gamma_{1}$ and $\gamma_{2}$, which exhibits robustness to different hyperparameter settings.

\subsection{Ablation Studies}
We perform ablation experiments, which is very important to understand the role of each component of MST-GAT. We gradually remove different modules to observe the changes in performance.
First, we remove the intra- and inter-modal attention modules in M-GAT. Secondly, we further remove the temporal convolution in MST-GAT.
Thirdly, to study the necessity of graph structure learning, we substitute the dynamic sparse graph implemented by TopK with a complete graph. In a complete graph, all nodes are connected to each other.
Finally, we discard the attention mechanism in the multi-head attention module and aggregate information by assigning equal weight to each neighbor.

\begin{table}[!t]
    \centering
    \footnotesize
    \caption{Performance comparison in terms of precision ({\%}), recall ({\%}), and F1-score ({\%}) of MST-GAT and its variants. The best results are highlighted in bold.}
    \scalebox{0.9}{
    \setlength{\tabcolsep}{8mm}{
    \begin{tabular}{lccc}
        \toprule
         \textbf{Method} & $\mathbf{\text{Prec}}$ & $\mathbf{\text{Rec}}$ & $\mathbf{\text{F1}}$ \\ \midrule
         \textsc{MST-GAT} & \textbf{98.73} & \textbf{72.41} & \textbf{83.55} \\
         - \textsc{Modal}              & 97.36 & 70.12 & 81.52 \\
         \ \ \ - \textsc{Temp}         & 96.73 & 69.54 & 80.91 \\
         \ \ \ \ \ \ - \textsc{TopK}    & 91.62 & 65.10 & 76.12 \\
         \ \ \ \ \ \ \ \ \ - \textsc{Att}  & 70.21 & 67.76 & 68.96 \\
         \bottomrule
    \end{tabular}}}
    \label{tab:ablation}
\end{table}

The results of MST-GAT and its variants on SWaT dataset are summarized in Table~\ref{tab:ablation}, and we find:
\textbf{\textit{(\romannumeral1)}} In the experiments, removing intra- and inter-modal attention modules results in significant performance degradation, which indicates that the explicit capture of intra- and inter-modal dependencies in multimodal time series is beneficial to improving performance.
We conjecture that the multimodal graph attention network is beneficial to obtain a better feature representation for MTS anomaly detection.
\textbf{\textit{(\romannumeral2)}} MST-GAT equipped with temporal convolution outperforms the model without it, which shows the necessity of modeling the temporal dependence in multimodal time series.
\textbf{\textit{(\romannumeral3)}}
The variant of MST-GAT that does not use the attention mechanisms performs worst than the other variants.
Since each univariate time series has very different properties, assigning the same weight to each neighbor introduces additional noise and cannot model the complex dependencies in multimodal time series.
\textbf{\textit{(\romannumeral4)}}
We can observe that the removal of each component consistently reduces performance, which proves the rationality of each component in MST-GAT.

\section{Conclusions}
\label{sec:cls}
In this paper, we have proposed a novel multimodal spatial-temporal graph attention network, termed MST-GAT, for multimodal time series anomaly detection. MST-GAT leverages the multimodal graph attention network and the temporal convolution network to capture the spatial correlation and temporal dependence among multimodal time series. The proposed model takes advantage of the reconstruction and prediction modules through joint training. Moreover, we propose an efficient anomaly interpretation approach for detected anomalies based on the reconstruction probability and the prediction value. Experimental results on benchmark datasets show that MST-GAT outperforms the state-of-the-art baselines and is able to provide interpretable results that are consistent with human intuition.
{
In the future, it makes sense to extend our framework to unaligned multimodal time series data, e.g., multimodal data collected by sensors of self-driving cars using different sampling rates.
Moreover, optimizing the memory footprint and execution time of the model to meet the needs of real-world deployments is an area worthy of future research.
}

\section*{Acknowledgements}

This work was supported by the NSFC Projects 62006078 and 62076096, the Shanghai Municipal Project 20511100900, the Shanghai Knowledge Service Platform Project (No. ZF1213), the Open Research Fund of KLATASDS-MOE, and the Fundamental Research Funds for the Central Universities.

\bibliography{ref}

\end{document}